\documentclass[journal,11pt,final,twocolumn]{IEEEtran}
\usepackage{algorithmic}
\usepackage{algorithm}
\usepackage{amsmath}
\usepackage{multirow}
\usepackage{amssymb}
\usepackage{graphicx}
\usepackage{diagbox}
\usepackage{color}
\usepackage{url}
\title{\LARGE \bf New Results on Multi-Step Traffic Flow Prediction}

\author{Arief~Koesdwiady and~Fakhri~Karray,~\IEEEmembership{Senior~Member,~IEEE}
\thanks{Arief Koesdwiady and Fakhri Karray are with the Centre for Pattern Analysis and Machine Intelligence, University of Waterloo, ON, Canada (corresponding author e-mail: abkoesdw@uwaterloo.ca)}}
\begin{document}
\maketitle
\thispagestyle{empty}
\pagestyle{empty}

%%%%%%%%%%%%%%%%%%%%%%%%%%%%%%%%%%%%%%%%%%%%%%%%%%%%%%%%%%%%%%%%%%%%%%%%%%%%%%%%
\begin{abstract}
In its simplest form, the traffic flow prediction problem is restricted to predicting a single time-step into the future. Multi-step traffic flow prediction extends this set-up to the case where predicting multiple time-steps into the future based on some finite history is of interest. This problem is significantly more difficult than its single-step variant and is known to suffer from degradation in predictions as the time step increases. In this paper, two approaches to improve multi-step traffic flow prediction performance in recursive and multi-output settings are introduced. In particular, a model that allows recursive prediction approaches to take into account the temporal context in term of time-step index when making predictions is introduced. In addition, a conditional generative adversarial network-based data augmentation method is proposed to improve prediction performance in the multi-output setting. The experiments on a real-world traffic flow dataset show that the two methods improve on multi-step traffic flow prediction in recursive and multi-output settings, respectively.
\end{abstract}

\section{Introduction}
Accurate and timely traffic flow prediction is essential for traffic management and allows travelers to make better-informed travel decisions. In some applications, it is often necessary to predict traffic flow not only accurately but also several steps ahead in the future. For example, in order for the traffic patch to manage a congested road and develop contingency plans, traffic dispatch may need to estimate traffic conditions at least 30 minutes in advance. However, most traffic flow prediction approaches were developed with single-step prediction methods. The multi-step prediction problem is significantly more difficult than its single-step variant and is known to suffer from degradation in predictions the farther we go in future time-steps. Therefore, it is essential to develop multi-step prediction approaches to achieve accurate multi-step traffic flow prediction.

Multi-step time series prediction tasks are defined as tasks of predicting the next $k$ values $[y_t, y_{t+1}, \ldots, y_{t+q}]$ given a historical time series $[y_t, y_{t-1}, \ldots, y_{t-p}]$, where $p, q > 1$ denote the past and future horizons. Generally, there are three main strategies for multi-step time series prediction: recursive, direct, and multi output~\cite{bontempi2012machine}. In the recursive strategy, a one-step model is first trained to fit the following function:

\begin{align}
y_{t+1} = f(y_t, \ldots, y_{t-p})
\end{align}

The learned model, $f$, predicts a multi-step time-series trajectory by repeatedly passing its predictions at one time step as input to the next time step. In the simple case where both the history and predictions are length-$1$ scalars, given $x(0)$, a model $f$ predicts $\hat{x}(1)$ as $f(x(0))$, $\hat{x}(2)$ as $f(\hat{x}(1))$, and so on\footnote{This simple case in presenting recursive prediction is used in this paper, but the ideas discussed apply to the general case.}. Due to accumulating errors and shifting input distribution, model predictions farther in the future increasingly drift from ground truth trajectories \cite{venkatraman2015improving}. Moreover, there is a mismatch between what the model is optimized for, i.e., single-step error, and what it is used for, i.e., multi-step prediction, that gives rise to optimistic error estimates during training \cite{taieb2014machine}. These weaknesses are present when the true single-step model is not identified during training \cite{taieb2016bias}, which is almost always the case in non-linear problems. Recent work showed that a learned model can be tuned to learn corrections to the drift patterns seen in the training data \cite{venkatraman2015improving}. This is an iterative training process, in which the training set is repeatedly augmented with additional data points of the form $(\hat{x}(t), x(t+1))$, where $\hat{x}(t)$ represents predictions of the model when applied to the training set. When the model is applied recursively to the training set points, it generates prediction trajectories of some length. The intuition of this iterative training process is that by augmenting the training set with samples of these predicted trajectories, coupled with the next-step ground truth values, the model can learn to correct the drift patterns in its predicted trajectories.

Alternatively, one can do without the recursive process by learning a separate model to directly predict each time-step in the future, i.e., the direct strategy, which is given as
\begin{align}
y_{t+h} = f_h(y_t, \ldots, y_{t-p})
\end{align}
where $h\in{1, \ldots, H}$ is the predictions horizon. In this strategy, multi-step predictions are obtained by concatenating the $H$ predictions. Unlike the recursive strategy, the direct strategy does not suffer from accumulating errors since it does not use any predicted values for the subsequent predictions. However, there are two major weaknesses possessed by this strategy. First, since each model is learned independently, dependencies between two distant horizons are not modeled. Second, this strategy requires large computational resources, i.e., time and space, since the number of models depends on the size of the prediction horizon.

\begin{figure}
  \centering
  \includegraphics[width=0.45\textwidth]{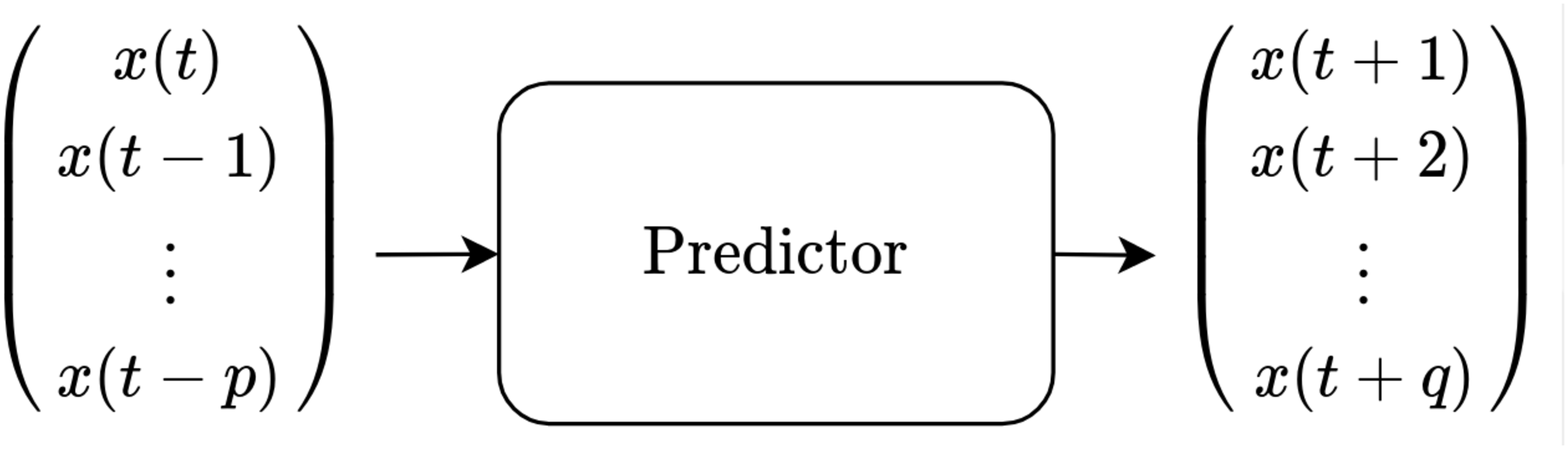}
  \caption{Multi-output strategy.}
  \label{fig:multi}
\end{figure}

The third strategy is the multi-output strategy. This strategy is defined as the problem of finding a model $f$ that predicts the future $[x(t+1), x(t+2), \cdots, x(t+q)]^\top$ given the historical data $[x(t), x(t-1), \cdots, x(t-p)]^\top$. The strategy requires a model that is able to produce multi-step predictions simultaneously, as depicted in Figure~\ref{fig:multi}. This way, each prediction uses the actual observations rather than the predicted ones. Therefore, accumulated errors are not of concern in this strategy. Moreover, this strategy can learn the dependency between inputs and outputs as well as among outputs. Hence, the strategy involves more complex models than the recursive one does, which directly translates to a slower training process and requires more training data to avoid over-fitting. While in some cases the direct and multi-output strategies can avoid some of the pitfalls of the recursive strategy, these models still suffer from degrading performance in the farther time-steps. Intuitively, there is a higher uncertainty associated with the farther future that makes it more difficult to forecast. Moreover, direct models can suffer from higher variance \cite{marcellino2006comparison}. Researchers have attempted to analyze theoretically and empirically the differences between recursive and direct/multi-output approaches and understand which would be more appropriate for a given problem~\cite{taieb2012review}, but the results of this effort so far have been inconclusive. In practice, all approaches continue to suffer from increasingly drifting predictions for farther time-steps.

Recently, an approach to counter the drifts in trajectories of multi-step predictions called Data as Demonstrator (DaD) is proposed in~\cite{venkatraman2015improving}, specifically in the context of recursive prediction models. The underlying idea in their approach is to use the drift patterns seen when a trained model is applied to the training data to tune that model such that it can compensate for these drifts. Another way to look at it is as a data augmentation technique that alleviates the mismatch between training and testing distributions. Inspired by this, two approaches to enhance multi-step prediction accuracy are introduced in this thesis. In the context of recursive models, a time-step-augmented model that implicitly learns to associate a different corrective action with different future time-steps is proposed. The model is an extension of the approach proposed in \cite{venkatraman2015improving}. This is also related to the \textit{Rectify} method proposed in~\cite{taieb2012recursive}, where a direct model is trained to correct the predictions of a recursive model at each time-step in the prediction trajectory. In the second approach, a data augmentation method that enhances multi-step prediction accuracy in multi-output models is proposed. Here, a conditional generative adversarial network (C-GAN) is used to learn a generator model that can mimic the historical patterns corresponding to the future patterns seen in the training data. Subsequently, this model is used to generate new history-future pairs that are aggregated with the original training data. 

The main contributions in this work are summarized as follows:
\begin{itemize}
	\item An extension to the algorithm presented in~\cite{venkatraman2015improving}, where information about the current time-step prediction is augmented in the model. This extension is called Conditional-DaD (C-DaD).
	\item A novel approach of generating new history-future pairs of data that are aggregated with the original training data using C-GAN.
	\item Comprehensive traffic flow prediction experiments involving recursive, direct, and multi-output strategies. In the recursive strategy, the vanilla approach, DaD, and C-DaD are experimented. Furthermore, the vanilla direct strategy and its modification using recursive strategy, namely \textit{Hybrid} method, are also presented. Finally, the proposed method C-GAN is compared with noise-augmented training strategy as well as the vanilla multi-output strategy.
\end{itemize}

The rest of this paper is structured as follows. Section \ref{meth} introduces the two proposed approaches. Section \ref{exp} describes the experimental setup and the data sets used to evaluate the proposed models. In Section \ref{res}, the experimental results are presented and discussed. Finally, the paper is concluded with some observations in Section \ref{conc}.

\section{Methodology}
\label{meth}
In this section, two methods to improve multi-step time-series prediction are introduced. An approach to improving recursive multi-step prediction, called Conditional-DaD (C-DaD), followed by a conditional-GAN-based data augmentation approach to improving multi-output multi-step prediction are introduced.
\subsection{Conditional-DaD}
One weakness of the approach presented in~\cite{venkatraman2015improving}, and recursive prediction generally, is that it does not take into consideration the number of steps predicted by the model so far. The amount of correction the model needs to add differs from time-step to another along a multi-step prediction trajectory since the deviation from the ground truth is less acute in early steps. Therefore, the model stands to benefit from having information about the current time-step along the prediction trajectory. In particular, the amount of correction the model needs to add is affected by the number of time-steps that have passed in which the model used its prediction as input to the next time-step.

\begin{figure}
	\centering
	\includegraphics[width=0.45\textwidth]{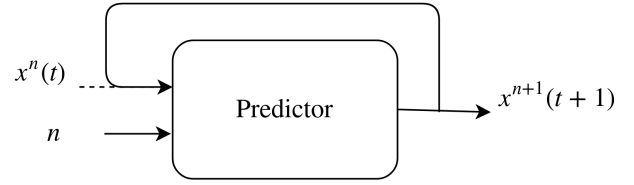}
	\caption{C-DaD recursive prediction model.}
	\label{C-DaD-figure}
\end{figure}

Based on this, an extension to DaD called C-DaD, in which the input is augmented with a representation of the current time-step, is proposed. For length-$1$, scalar history, $x(t)$, the single-step prediction model is modified to accept an augmented vector, $[x^n(t), v_n]^\top$, where $n$ is the prediction time-step, and $v_n$ is a representation of $n$. In the presentation and experiments, $v_n = n$ is used. An illustration of this is shown in Fig. \ref{C-DaD-figure}.

A C-DaD model learns a single mapping that is a function of the number of predicted values that have been recycled as input (and also a function of the time-series history). This arrangement allows the model to output different $x(t+1)$ for the same $x(t)$ input, depending on the current time-step along the prediction trajectory, and, hence, allows the model to learn different corrections for different time-steps. This setup differs from the parameterized recursive prediction approach, where a different set of parameters are learned for each time-step in the future.

Training a C-DaD model follows a similar process to the meta-algorithm proposed in \cite{venkatraman2015improving}, the difference being in the addition of the time-step representation. Algorithm \ref{cdad-algorithm} describes this process. In short, the time-step-augmented training data is generated by forward-passing the original training data through a base model. Next, a C-DaD model is iteratively trained, and a new augmented training data set is generated every epoch by passing the original data through the previous C-DaD model. Furthermore, the final model is selected based on the performances of all models on the validation data set.

\begin{algorithm}
	\centering
	\caption{C-DaD training process. \label{cdad-algorithm}}
	\begin{algorithmic}[1]
		\renewcommand{\algorithmicrequire}{\textbf{Initialize:}}
		\renewcommand{\algorithmicensure}{\textbf{Input:}}
		\ENSURE~~\\
			\textbullet~$X$---time-series points, $x(t)$.\\
			\textbullet~\emph{predict\_recursively}---procedure that takes as input a sequence of points, $X$, an integer, $N$, and a single-step prediction model, \emph{net}, and outputs the length-$N$ recursive predictions of \emph{net} applied to the data points in $X$. The output of this is of the form $\hat{x}^n(t)$, where $n$ indicates the recursion index.\\
			\textbullet~\emph{predict\_recursively\_aug}---similar procedure to \emph{predict\_recursively}, except that the model is applied to time-step-augmented inputs $X_{aug}$ of the form $[x^n(t), n]^\top$.

		\REQUIRE~~\\
			\textbullet~$N$---number of recursive time steps to use.\\
			\textbullet~epochs---number of iterations to train the model.\\
		\STATE Use $X$ to build a training set $D$ of the form $(x(t), x(t+1))$.\\
		\STATE Initialize single-step-prediction base model, $M_{base}$, and train it using $D$.\\
		\STATE $\hat{X}^{(N)} \leftarrow $ \emph{predict\_recursively}($X, N, M_{base}$)\\
		\STATE Use $X$ and $\hat{X}^{(N)}$ to build a training set $D_{aug}$ of the form $([\hat{x}^n(t), n]^\top, x(t+1))$, $n \in [0, N]$\\
		\STATE Initialize single-step-prediction augmented model, $M_0$.\\
		\FOR {$n= 1,\ldots, K$}
			\STATE Build a time-step-augmented inputs $X_{aug}$
			\STATE $\hat{X}^{(N)} \leftarrow  predict\_recursively\_aug(X_{aug}, N, M_{n-1})$
			\STATE Use $X_{aug}$ and $\hat{X}^{(N)}$ to build a training set $D_{aug}$
			\STATE $M_n = train(D_{aug})$
		\ENDFOR
		\RETURN $M_n$ with lowest error on the validation data.
	\end{algorithmic}
\end{algorithm}
\subsection{Conditional-GAN Data Augmentation}
In some applications, recursive models do not perform well compared to other approaches such as multi-output models \cite{marcellino2006comparison}. Nonetheless, the performance of multi-output models suffers degradation as the prediction time-step increases. In this section, a C-GAN-based data augmentation approach to improve  multi-output multi-step time series prediction is introduced.

The multi-output strategy requires a model that is able to produce multi-step predictions simultaneously. This way, each prediction uses the actual observations rather than the predicted ones. Therefore, accumulated errors are not of concern in this strategy. Moreover, this strategy can learn the dependency between inputs and outputs as well as among outputs. Hence, the strategy involves more complex models than the recursive one does, which directly translates to a slower training process and requires more training data to avoid over-fitting.

Similar to the recursive strategy, a data augmentation method can be applied to improve the multi-step time series prediction. One simple way to augment the data is to contaminate the features with noise and pair them with the actual labels. This method can increase the multi-step prediction performance if the noise is carefully chosen. Poor choice of noise, however, may significantly degrade the prediction performance. In this work, an alternative method, i.e., Generative Adversarial Network (GAN)~\cite{goodfellow2014generative}, to augment the data by learning a distribution over input conditioned on the output is proposed.

GAN is a framework for estimating a distribution in an adversarial manner. It simultaneously trains two models, namely a generative model $G$ and discriminative model $D$. The discriminative model is trained to maximize the probability of assigning appropriate labels for both samples coming from the training data and generative model. Simultaneously, the generative model is trained to minimize $\log (1-D(G(z))$, where $z$ is a random sample from an input noise distribution. Both $D$ and $G$ are playing a two-player minimax game with the value function $V(G, D)$ given as follows:
\begin{align}
\min_G\max_D V(D,G)= & \mathbb{E}_{x\sim p_{data}(x)}\left[ \log D(x)\right]\nonumber\\ & + \mathbb{E}_{z\sim p_z(z)}\left[\log(1-D(G(z)))\right]
\end{align}

\begin{figure}[!htbp]
	\centering
	\includegraphics[width=0.45\textwidth]{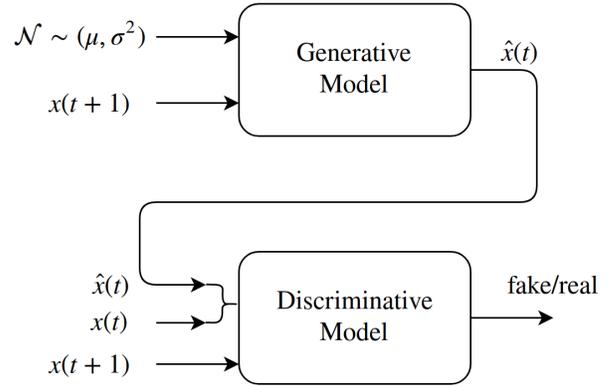}
	\caption{C-GAN for generating time series data.}
	\label{fig:gan}
\end{figure}

An extension of GAN, namely conditional generative adversarial nets (C-GAN), is proposed in~\cite{mirza2014conditional}. In this extension, both $G$ and $D$ are conditioned on some extra information, which can be any kind of auxiliary information such as class labels. There have been some research applying C-GAN to discrete labels~\cite{denton2015deep}, text~\cite{reed2016generative}, and images~\cite{yoo2016pixel}. In this work, C-GAN is trained to generate inputs (historical data) given the actual labels (future data). In both the generative and discriminative models, the actual labels are concatenated with noise. This idea is illustrated in Figure~\ref{fig:gan}, where $x(t+1)$ is the label (future data), $\hat x(t)$ is the generated inputs (past data). The pair of generated inputs and actual labels then are augmented in the original training data for multi-output time series prediction.

% From the structure of GAN, it seems that we can create a generative model using historical data as the input and future data as the output. This scenario lets us to use the generative model as the multi-output predictor, which can be done if we do not concatenate the noise with the input. However, the generative model becomes deterministic. In other words, the generative model is not really a generative model anymore.
Using this method, an infinite amount of data to enhance the predictor performance can be generated. In addition, using the generated data, there is no need any special treatments in the training process. It can be done in a standard multi-output training without any iterative training processes. 
\section{Datasets and Experimental Settings}
\label{exp}
To test the performance of the proposed methods, we conduct experiments using a traffic flow data set. The data set was downloaded from the Caltrans Performance Measurements Systems (PeMS)~\cite{PeMS}. The original traffic flow was sampled every 30 seconds. These data were aggregated into 5-min duration by PeMS. Highway Capacity Manual 2010~\cite{manual2010volumes} recommends to aggregate the data further into 15-min duration. We collected the traffic flow data of a freeway from January 1\textsuperscript{st} 2011 to December 31\textsuperscript{st} 2012. We use data from from January 1\textsuperscript{st} 2011-from August 31\textsuperscript{st} 2011 for training, September 1\textsuperscript{st} 2011-December 31\textsuperscript{st} 2011 for validation, and the rest for testing.

Three sets of experiments are conducted using the data set. In the first set, the vanilla recursive strategy, DaD, and C-DaD are implemented on the dataset. The main goal of these experiments is to investigate which strategy performs better in the recursive setting. Next, the vanilla direct strategy and Hybrid approach are applied to the dataset. Subsequently, the proposed C-GAN data augmentation is applied and compared with the vanilla multi-output strategy and noise data augmentation model. The number of the time steps for the multi-step prediction is chosen to be 8. Furthermore, the performances of the best models from of each strategy are compared and analyzed. The performance of each of the experiments is evaluated using Mean Squared Error (MSE) and Mean Absolute Error (MAE). To illustrate the superiority of the proposed methods, the percentages of improvement of the errors with respect to the baselines are computed. 
\section{Results and Analysis}
\label{res}
In the first set of the experiments, three recursive approaches are applied to the dataset. Each approach uses similar base predictor, which is a deep neural network (DNN). To have fair comparisons, the configurations of the DNN for all of the approaches, in terms of the number of hidden layers, number of hidden units, activation function, and tricks used for the training, are kept identical. The main difference is that in the C-DaD approach the input is augmented with the time-step information, which means there are extra weights associated with this input.

The base DNN is configured to have 2 hidden layers where each layer contains 150 hidden units, and the activation function is selected to be ReLU. Since it is a regression problem, a linear activation is used in the output layer with MSE as the loss function. Furthermore, the data are min-max normalized between 0 and 1. Moreover, to avoid the model from over-fitting too easily, drop-out regularizers with the rate equals to 0.1 are implemented on each layer. In addition, the Adam~\cite{kingma2014adam} algorithm is used for the gradient-based optimization.

The summary of the performance of the recursive approaches can be seen in Table~\ref{recursive}. The table shows that, with respect to the vanilla recursive approach, the DaD and C-DaD approaches have successfully improved the overall MSE and MAE. This shows that reusing the prediction results as the input data, together with the original training data, leads to improvement in the performances. Furthermore, augmenting the information of the time step in the model, i.e., C-DaD approach, can further improve the performances for more than 2 times in the MSE and almost 1.5 times in the MAE, as can be seen in the table. These performances are achieved after 25 and 29 iterations in the C-DaD and DaD approaches, respectively.

\begin{table}
	\caption {Summary of recursive multi-step prediction performances.\label{recursive}}
	\centering
	\begin{tabular}{l|c|c|c|c}
		\hline
		\hline
		\backslashbox{Models}{Perf.}& MSE & \% Improv. & MAE & \% Improv.\\
		\hline
		Recursive & 0.0101 & - & 0.0781 & - \\
		DaD & 0.0092 & 8.16 & 0.0627 & 19.64\\
		C-DaD & 0.0078 & 22.92 & 0.0563 & 27.89\\
		\hline
		\hline
	\end{tabular}
\end{table}

\begin{figure}
	\centering
	\includegraphics[width=0.5\textwidth]{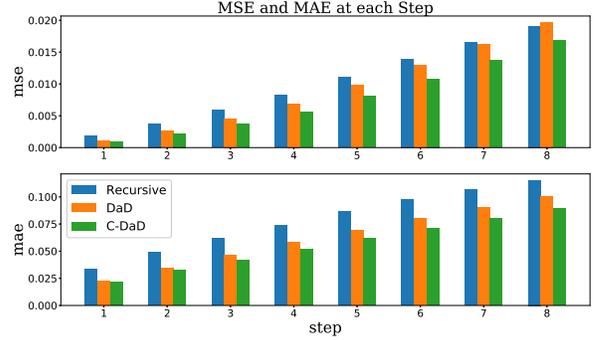}
	\caption{MSE (top) and MAE (bottom) at each step for the recursive approaches.}
	\label{fig:recursive_step}
\end{figure}

Figure~\ref{fig:recursive_step} depicts the error occurs at each time step. It shows that, in the early step, both the DaD and C-DaD approaches perform significantly better than the vanilla recursive does. However, at time step equals to 8, the MSE of the DaD approach is worse than that of the vanilla recursive, while the C-DaD approach is able to maintain its performances all the way through the last step. This is not the case for the MAE, where both the DaD and C-DaD approaches are able to maintain its performances at all time steps.

\begin{figure}
	\centering
	\includegraphics[width=0.5\textwidth]{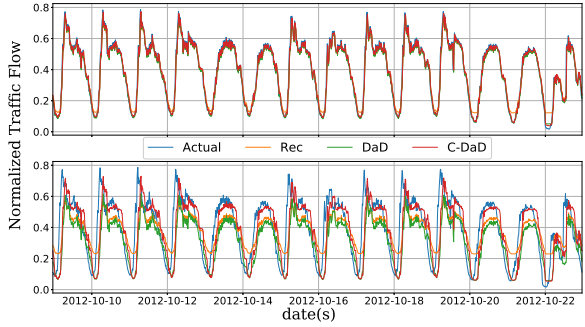}
	\caption{Recursive approaches traffic flow prediction results at time step 1 (top) and 8 (bottom).}
	\label{fig:traffic_rec}
\end{figure}

The results of the traffic flow predictions for the recursive approaches can be seen in Figure~\ref{fig:traffic_rec}. This figure presents traffic flow predictions at time step 1 and 8. At time step 1, all the approaches produce similar predictions, which are very close to the actual traffic flow. However, at time 8, only the C-DaD approach is showing an acceptable prediction. Indeed, the augmentation of the time-step information provides an extra dimension to the model, which helps the model to understand the state of the prediction and learn better. It can be concluded that adding this extra dimension is worth the effort.

In the second set of experiments, two direct approaches, namely vanilla direct and Hybrid approaches, are tested. Similar base predictors as the previous set of experiments are used. The number of models trained in this approach depends on the size of the future horizon. Since the number of time steps is 8, then the number of models in each approach will be 8. The main difference between the vanilla direct and Hybrid approaches is in the size of the input. The number of input in the Hybrid increases as the time step increases while the number of input in the vanilla direct is static.

\begin{table}
	\caption {Summary of direct multi-step prediction performances.\label{direct}}
	\centering
	\begin{tabular}{l|c|c|c|c}
		\hline
		\hline
		\backslashbox{Models}{Perf.}& MSE & \% Improv. & MAE & \% Improv.\\
		\hline
		Direct & 0.0090 & - & 0.0715 & - \\
		Hybrid & 0.0082 & 8.68 & 0.0674 & 5.63\\
		\hline
		\hline
	\end{tabular}
\end{table}

The performance of the direct approaches is summarized in Table~\ref{direct}. In this table, it can be seen that the vanilla direct approach has smaller errors than the vanilla recursive does. This is expected since in the vanilla direct approach the accumulating error problem does not exist. Each time-step prediction is handled independently by each model. Furthermore, the Hybrid approach improves the vanilla direct performance considerably. However, if it is compared with the best performance in the recursive approach, C-DaD is still shown its superiority. This may be attributed to the accumulating errors when the previous predictions are used as inputs in the subsequent models.

\begin{figure}
	\centering
	\includegraphics[width=0.5\textwidth]{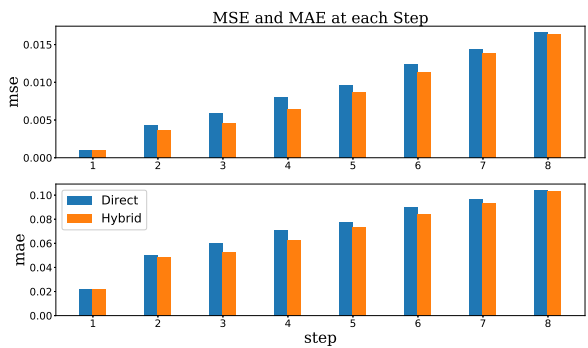}
	\caption{MSE (top) and MAE (bottom) at each step for the direct approaches.}
	\label{fig:direct_step}
\end{figure}

From Figure~\ref{fig:direct_step}, it can be seen that the errors in the first step are identical. This is possible because the models in this step are practically the same since the previous predictions are not utilized yet. Overall, the Hybrid approach improves the prediction errors in all time steps. However, a closer look suggests that the Hybrid model performances degrade as the time step increases. Indeed, at the later steps, the accumulating errors dominate the input of the model. In this type of approach, the performance in the next step highly depends on the one in the previous model.

\begin{figure}
	\centering
	\includegraphics[width=0.5\textwidth]{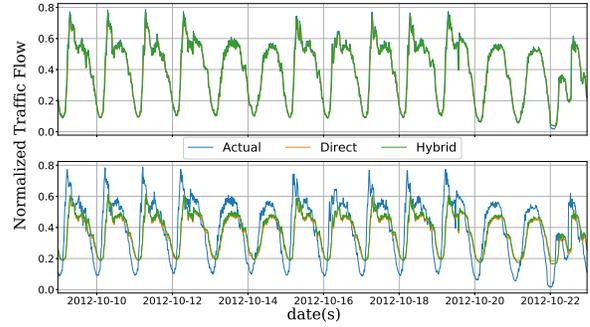}
	\caption{Direct approaches traffic flow prediction results at time step 1 (top) and 8 (bottom).}
	\label{fig:traffic_direct}
\end{figure}

Figure~\ref{fig:traffic_direct} shows that, initially, the Hybrid approach produces acceptable prediction, which is not the case in the last step. This is reciprocal with the errors shown in Figure~\ref{fig:direct_step}, where there is almost no improvement in MSE and MAE achieved in the last step. In comparison with C-DaD, this method is computationally inefficient since it requires several models for multi-step predictions. With 8 times less computational efforts, the C-DaD approaches perform considerably well than the Hybrid method does.

\begin{figure*}
	\centering
	\includegraphics[width=0.75\textwidth]{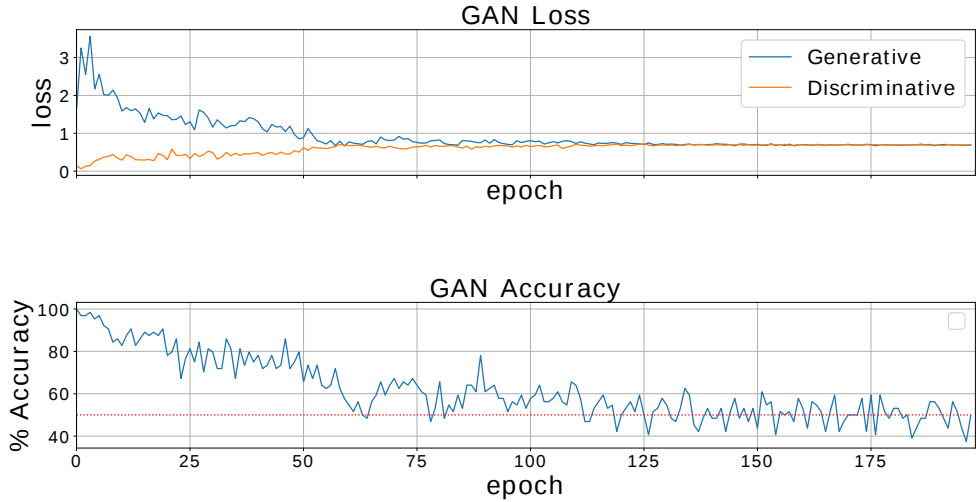}
	\caption{GAN loss and accuracy progressions.}
	\label{fig:gan_loss}
\end{figure*}

The last set of experiments involves three multi-output approaches: vanilla multi, noise-augmented, and C-GAN approaches. The base DNN is configured to have 2 hidden layers where each layer contains 150 hidden units, and the activation function is selected to be ReLU. The three approaches use identical models, including the output layer size. In the noise-augmented approach, the data are contaminated with a Gaussian noise with mean equals to 0 and variance equals to 0.1. After several trials, this variance is found to produce the best performance on the validation data set. Meanwhile, the discriminative and generative models of the C-GAN use DNN with a similar configuration as the base DNN. The important aspect of training the C-GAN is the learning rates of both the discriminative and generative models. Usually, the discriminative model is configured to learn faster than the generative model. This way the discriminative loss stays low, which makes it stays ahead of discriminating new strange representations from the generative model. The evolution of the losses in the DC-GAN is depicted in Figure~\ref{fig:gan_loss}. It can be seen that the losses of both the discriminative and the generative models converge. Furthermore, the C-GAN accuracy converges to 50\%, which means the discriminator is not able to distinguish the data generated by the generative model from the actual data. Therefore, it can be concluded that the generative model acts as a distribution that mimics the training data.

The overall performances of the multi-output traffic flow prediction approaches are summarized in Table~\ref{multi}. So far, the lowest MSE using the vanilla approaches is obtained by the multi-output approach. It is attributed to the fact that in the multi-output setting, the accumulating errors problem does not exist and the dependencies between time steps are modeled. In the noise-augmented approach, a poor choice of noise may significantly degrade the prediction performances. However, in this experiment, the noise has been successfully chosen as it is evident in the prediction performances improvements. Furthermore, the best improvement is achieved when the original data is augmented with the one generated by the generative model. Therefore, C-GAN can be seen as an intelligent way for data augmentation.

\begin{table}
	\caption {Summary of multi-output multi-step prediction performances.\label{multi}}
	\centering
	\begin{tabular}{l|c|c|c|c}
		\hline
		\hline
		\backslashbox{Models}{Perf.}& MSE & \% Improv. & MAE & \% Improv.\\
		\hline
		Multi & 0.0089 & - & 0.0718 & - \\
		Noise & 0.0082 & 8.13 & 0.0671 & 6.57\\
		GAN & 0.0072 & 18.47 & 0.0576 & 19.71\\
		\hline
		\hline
	\end{tabular}
\end{table}

\begin{figure}
	\centering
	\includegraphics[width=0.5\textwidth]{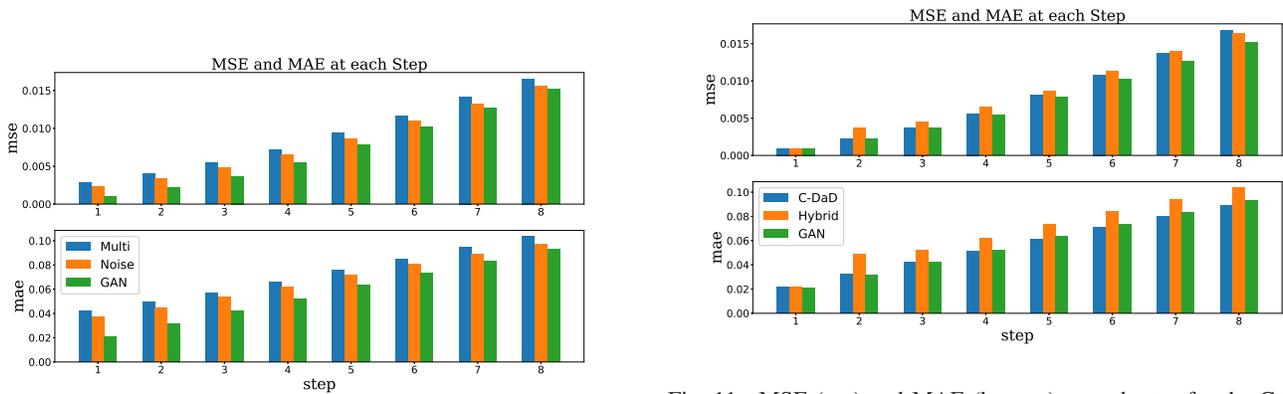}
	\caption{MSE (top) and MAE (bottom) at each step for the multi approaches.}
	\label{fig:multi_step}
\end{figure}

\begin{figure}
	\centering
	\includegraphics[width=0.5\textwidth]{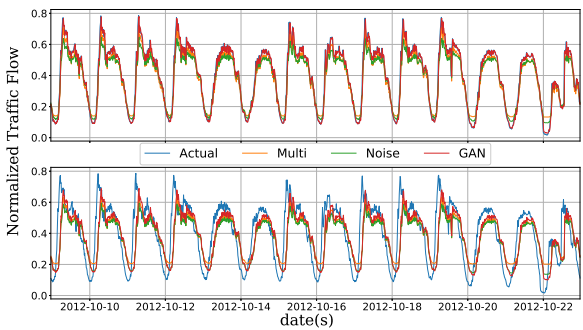}
	\caption{Multi approaches traffic flow prediction results at time step 1 (top) and 8 (bottom).}
	\label{fig:traffic_multi}
\end{figure}

Figure~\ref{fig:multi_step} depicts the MSE and MAE of the multi-output approaches at all time steps. Both the noise-augmented and C-GAN approaches consistently produce improved traffic flow predictions all the way through the all time steps. Furthermore, in Figure~\ref{fig:traffic_multi}, it can be seen that the performances of the vanilla multi and noise-augmented approaches are poor in the first time step. Indeed, learning several time-steps simultaneously is more difficult than learning 1 step only as it is done in the recursive and direct approaches. However, the proposed C-GAN approach is able to significantly improve the early time step predictions and overall performances.

\begin{figure}
	\centering
	\includegraphics[width=0.5\textwidth]{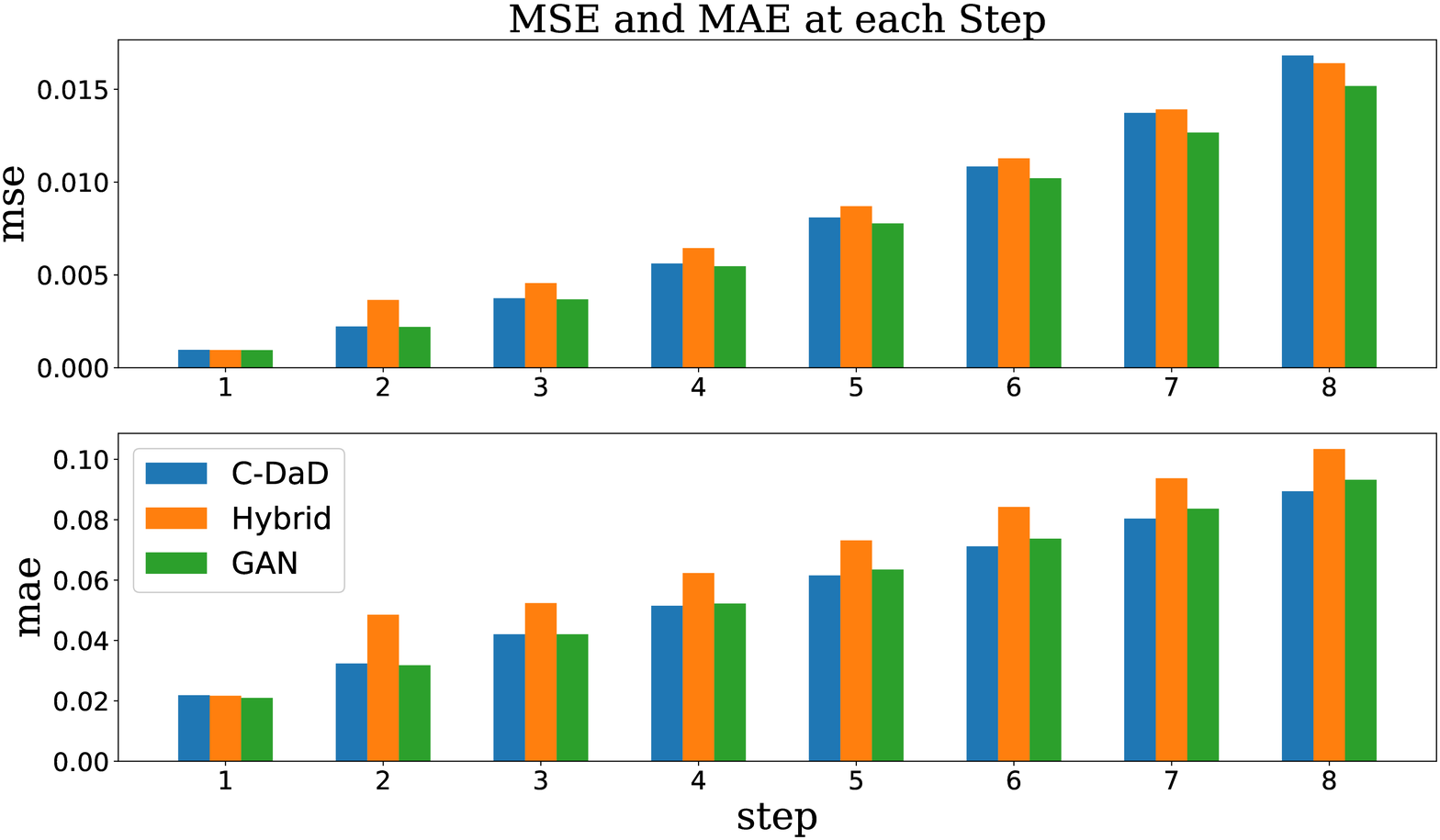}
	\caption{MSE (top) and MAE (bottom) at each step for the C-DaD, Hybrid, and C-GAN approaches.}
	\label{fig:all_step}
\end{figure}

Finally, the comparison of C-DaD, Hybrid, and C-GAN approaches is depicted in Figure~\ref{fig:all_step}. This figure shows that the C-GAN approach has shown its superiority in term of MSE at all time steps. However, in term of MAE, the C-DaD approach is better at the later steps compared to the C-GAN approach. Indeed, based on the prediction plots in~Figure\ref{fig:traffic_rec} and~Figure\ref{fig:traffic_multi}, it can be seen that the C-DaD approach produce better traffic flow prediction than the C-GAN does. In~\cite{willmott2005advantages}, it is suggested that MAE is more natural and quite often MSE can be misleading and is not a good indicator of average model performance because it is a function of two characteristics of a set of errors, rather than of one. In addition,~\cite{chai2014root} demonstrates that the MSE is more appropriate to represent model performance when the error distribution is expected to be Gaussian. 
\section{Conclusions}
This paper proposed two methods to improve multi-step traffic flow predictions: C-DaD and C-GAN approaches. The first approach is developed using recursive strategy and inspired by previous work~\cite{venkatraman2015improving}. This approach augments the information about the current time step and follows a similar training process to the meta-algorithm proposed in~\cite{venkatraman2015improving}. The second model is developed using multi-output strategy and utilizes the ability of GAN in mimicking a data set distribution. The C-GAN model is developed to generate historical data conditioned on the future data. This way, the original data set can be enriched with an infinite amount of historical-future pairs of data for training purposes.

The experiments show that the proposed approaches are able to improve multi-step traffic predictions relative to their vanilla approaches. Moreover, in term of MSE, the C-GAN approach performs better than the all of the approaches. However, in the latter steps, the MAE of the C-DaD is lower than all the experimented approaches. Compared to the C-DaD, the training of the C-GAN approach is fairly simpler since it does not require iterative training once the new data are generated. However, there are applications where it is more efficient to use recursive model, such as for video sequence prediction, and in such application recursive prediction can benefit from the improvement offered by the C-DaD approach. 
\label{conc}
\bibliography{reference} 
\bibliographystyle{ieeetr}
\begin{IEEEbiography}[{\includegraphics[width=1in,height=1.25in,clip,keepaspectratio]{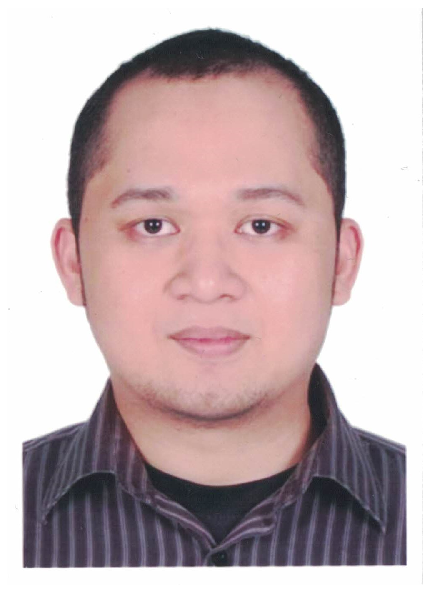}}]{Arief Koesdwiady}
	received the B.Eng degree in physics engineering from Institute Teknologi Bandung, Indonesia, and the M.Sc. degree in control system engineering from King Fahd University of Petroleum and Minerals, Dhahran, Saudi Arabia, in 2008 and 2013, respectively. He is currently pursuing the Ph.D. degree from the Centre for Pattern Analysis and Machine Intelligence at the University of Waterloo, Waterloo, ON, Canada. His current research interests include artificial intelligence, machine learning, deep learning, big data, intelligent transportation systems, and data fusion.
\end{IEEEbiography}
\vfill
\begin{IEEEbiography}[{\includegraphics[width=1in,height=1.25in,clip,keepaspectratio]{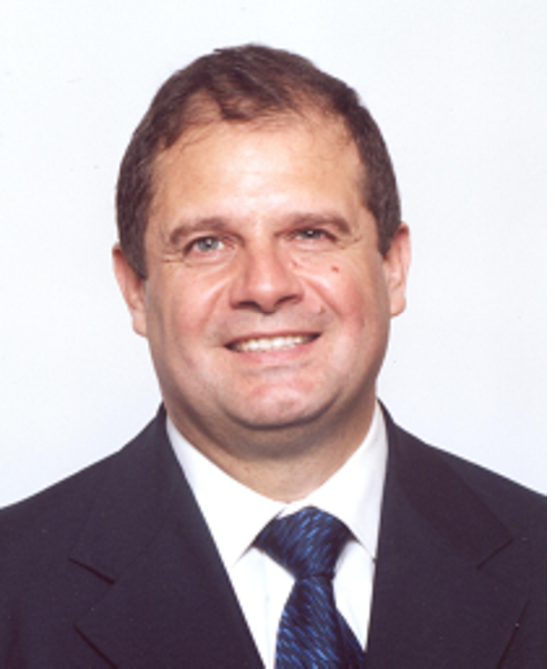}}]{Fakhreddine Karray} received the Dip-Ing. degree in electrical engineering from ENIT, Tunis, Tunisia, and the Ph.D. degree from the University of Illinois, Urbana Champaign, Champaign, IL, USA. He is the University Research Chair Professor in Electrical and Computer Engineering and Co-Director of the Center for Pattern Analysis and Machine Intelligence Center at the University of Waterloo, Waterloo, ON, Canada. His current research interests include intelligent systems, soft computing, sensor fusion, and context aware machines with applications to intelligent transportation systems, cognitive robotics, and natural man-machine interaction. He has co-authored over 400 technical articles, a textbook on soft computing and intelligent systems, six edited textbooks, and 20 textbook chapters. He holds 15 U.S. patents. Dr. Karray has Chaired/Co-Chaired several international conferences in his area of expertise and has served as a Keynote/Plenary Speaker on numerous occasions. He has also served as an Associate Editor/Guest Editor for a number of journals, including Information Fusion, the IEEE Trans. on CYBERNETICS, the IEEE Trans. on Neural Networks and Learning, the IEEE Trans. on Mechatronics, and the IEEE Computational Intelligence Magazine. He is the Chair of the IEEE Computational Intelligence Society (CIS) Chapter in Kitchener-Waterloo,
\end{IEEEbiography}
\vfill
\end{document}